\documentclass[reqno]{amsart}

\usepackage{amssymb}
\usepackage{graphicx}
\usepackage{hyperref}

\theoremstyle{remark}
\newtheorem{remark}{Remark}

\begin{document}

\title{A Minimal Six-Point Auto-Calibration Algorithm}

\author{E.V. Martyushev}

\date{July 15, 2013}


\address{South Ural State University, 76 Lenin Avenue, Chelyabinsk 454080, Russia}
\email{mev@susu.ac.ru}

\begin{abstract}
A non-iterative auto-calibration algorithm is presented. It deals with a minimal set of six scene points in three views taken by a camera with fixed but unknown intrinsic parameters. Calibration is based on the image correspondences only. The algorithm is implemented and validated on synthetic image data.
\end{abstract}

\maketitle

\section{Introduction}

The problem of camera calibration is a necessary part of computer vision applications such as path-planning and navigation for robots, self-parking systems, camera based industrial detection and recognition, etc. At present, a great deal of calibration algorithms and techniques have been developed. Some of them require to observe a planar pattern viewed at several different orientations~\cite{Heikkila, Zhang}. Other methods use the 3-dimensional calibration objects consisting of two or three pairwise orthogonal planes, whose geometry is known with good accuracy~\cite{Tsai}. In contrast with the just mentioned methods, the \emph{auto-calibration} does not require any special calibration objects~\cite{Hartley92, Hartley94, MF, MC, QT, Triggs}, so only point correspondences in several uncalibrated views are required. This provides the auto-calibration approach with a great flexibility and makes it indispensable in some real-time applications.

In this paper we give a new non-iterative solution to the auto-calibration problem in a minimal case of six scene points in three views, provided that the intrinsic parameters of a moving camera are fixed. Our method consists of two major steps. First, we use the efficient six-point three-view algorithm from~\cite{SZHT} to solve for projective reconstruction. Then, using the well-known constraints on the absolute dual quadric~\cite{HZ, Triggs}, we produce a system of non-linear polynomial equations, and resolve it in a numerically stable way by a series of Gauss-Jordan eliminations with partial pivoting.

The rest of the paper is organized as follows. In Section~2, we briefly recall how to construct a projective reconstruction from six matched scene points in three uncalibrated views. In Section~3, an algorithm of metric upgrading of the projective reconstruction is described. In Section~4, we test the algorithm on a set of synthetic data. Section~5 concludes.

\subsection{Notation}

We use $\mathbf a, \mathbf b, \ldots$ for column vectors, and $\mathbf A, \mathbf B, \ldots$ for matrices. For a matrix $\mathbf A$, the entries are $A_{ij}$ or $(\mathbf A)_{i, j}$, the transpose is $\mathbf A^{\mathrm T}$, and the determinant is $\det(\mathbf A)$. For two vectors $\mathbf a$ and $\mathbf b$, the vector product is $\mathbf a\times \mathbf b$, and the scalar product is $\mathbf a^{\mathrm T} \mathbf b$. We use $\mathbf I_n$ for identical matrix of size $n\times n$ and $\mathbf 0_n$ for zero $n$-vector.

\section{Projective Reconstruction}
\label{sec:proj}

First of all, to avoid any degeneracies, we restrict ourselves to the ``general position case'' both for scene points and camera motions, i.e., the sequence of camera motions is assumed to be non-critical and all the observed points do not lie on critical surfaces in a sense of~\cite{Sturm}. In particular, this means that the scene is non-planar and the motion is not a pure translation or rotation around the same axis.

Given three uncalibrated images of six points of a rigid scene, we first produce a projective reconstruction of the cameras applying the minimal 3-view algorithm from~\cite{SZHT}. Recall that the output of this algorithm is either one or three real solutions for the homogeneous coordinates of the sixth scene point~$\mathbf X_6$, whereas the first five points are chosen to be the vectors of standard basis of the projective 3-space. The twelve entries of the camera matrix~$\mathbf P_i$ are then recovered by solving the twelve linearly independent equations (for each $i = 1, 2, 3$):
\[
\mathbf x_{ij} \times \mathbf P_i \mathbf X_j = \mathbf 0_3, \quad j = 1, \ldots, 6,
\]
where $\mathbf x_{ij}$ is the image of $\mathbf X_j$ under the projection~$\mathbf P_i$. Thus we found
\[
\mathbf P_i = \begin{bmatrix}\mathbf A_i & \mathbf a_i\end{bmatrix}, \quad i = 1, 2, 3.
\]

Using the projective ambiguity~\cite{HZ}, we transform the obtained camera matrices to
\begin{equation}
\begin{split}
\label{eq:projective}
\mathbf P'_1 &= \mathbf P_1 \mathbf H_0 = \begin{bmatrix}\mathbf I_3 & \mathbf 0_3\end{bmatrix},\\
\mathbf P'_2 &= \mathbf P_2 \mathbf H_0 = \begin{bmatrix}\mathbf B_2 & \mathbf b_2\end{bmatrix},\\
\mathbf P'_3 &= \mathbf P_3 \mathbf H_0 = \begin{bmatrix}\mathbf B_3 & \mathbf b_3\end{bmatrix},
\end{split}
\end{equation}
where
\[
\mathbf H_0 = \begin{bmatrix}\mathbf A_1^{-1} & -\mathbf A_1^{-1}\mathbf a_1 \\ \mathbf 0_3^{\mathrm T} & 1\end{bmatrix}.
\]

\section{Metric Reconstruction}
\label{sec:metric}

The projective reconstruction~\eqref{eq:projective} is the starting point for our auto-calibration algorithm. Let the metric camera matrices be represented as
\begin{equation}
\begin{split}
\label{eq:metric}
\mathbf P_1^M &= \mathbf K\begin{bmatrix}\mathbf I_3 & \mathbf 0_3 \end{bmatrix},\\
\mathbf P_2^M &= \mathbf K\begin{bmatrix}\mathbf R_2 & \mathbf t_2 \end{bmatrix},\\
\mathbf P_3^M &= \mathbf K\begin{bmatrix}\mathbf R_3 & \mathbf t_3 \end{bmatrix},
\end{split}
\end{equation}
where $\mathbf R_i$ and $\mathbf t_i$ are the rotation matrix and translation vector respectively, and $\mathbf K$ is an upper triangular matrix called the \emph{calibration matrix} of the camera. It is assumed to be identical for all three views. Our goal is to estimate $\mathbf K$ and then upgrade the projective cameras to the metric ones.

Auto-calibration determines a $4\times 4$ projective matrix $\mathbf H$, that transforms the projective camera $\mathbf P'_i$ from~\eqref{eq:projective} into a metric camera~$\mathbf P_i^M$ from~\eqref{eq:metric}, i.e.,
\begin{equation}
\label{eq:update}
\mathbf P_i^M = \mathbf P'_i \mathbf H, \quad i = 1, 2, 3.
\end{equation}
The matrix $\mathbf H$ must have the form~\cite{HZ}:
\[
\mathbf H = \begin{bmatrix}\mathbf K & \mathbf 0_3 \\ -\mathbf p^{\mathrm T}\mathbf K & 1 \end{bmatrix}
\]
for some 3-vector $\mathbf p$. Then the entries of~$\mathbf H$ are constrained by~\cite{Faugeras, HZ}
\begin{equation}
\begin{split}
\label{eq:autocalib}
\lambda\boldsymbol{\omega}^* &= \mathbf P'_2\mathbf{Q}^*_\infty {\mathbf P'_2}^{\mathrm T},\\
\mu\boldsymbol{\omega}^* &= \mathbf P'_3\mathbf{Q}^*_\infty {\mathbf P'_3}^{\mathrm T},
\end{split}
\end{equation}
where $\boldsymbol{\omega}^* = \mathbf K \mathbf K^{\mathrm T}$ is the \emph{dual image of the absolute conic}, $\lambda$, $\mu$ are scalars and $4\times 4$ matrix
\[
\mathbf{Q}^*_\infty = \begin{bmatrix}\boldsymbol{\omega}^* & \mathbf q \\ \mathbf q^{\mathrm T} & r \end{bmatrix},
\]
with $\mathbf q = - \boldsymbol{\omega}^* \mathbf p$, $r = \mathbf p^{\mathrm T} \boldsymbol{\omega}^* \mathbf p$, is called the \emph{absolute dual quadric}~\cite{Triggs}.

Thus, constraints~\eqref{eq:autocalib} give $12$ equations in $11$ variables: $r$, $q_1$, $q_2$, $q_3$, five components of $\boldsymbol{\omega}^*$ (recall that $\omega^*_{33} = 1$), $\lambda$ and~$\mu$. Let us rewrite these equations in form
\begin{equation}
\label{eq:Cx}
\mathbf C\,\mathbf x = \mathbf 0_{12},
\end{equation}
where
\begin{equation}
\label{eq:matrixC}
\mathbf C = \mathbf C(\lambda, \mu) = \begin{bmatrix} \mathbf 0_{6\times 4} & \lambda\mathbf I_6 \\ \mathbf 0_{6\times 4} & \mu\mathbf I_6 \end{bmatrix} - \mathbf D,
\end{equation}
$\mathbf D$ is a $12\times 10$ scalar matrix, and
\[
\mathbf x = \begin{bmatrix} r & q_1 & q_2 & q_3 & \omega^*_{11} & \omega^*_{12} & \omega^*_{13} & \omega^*_{22} & \omega^*_{23} & 1\end{bmatrix}^{\mathrm T}
\]
is a monomial vector.

It follows that the determinant of any $10\times 10$ submatrix of $\mathbf C$ must vanish. Denote by $S_i(\lambda, \mu)$ the determinant of a submatrix of $\mathbf C$ obtained by eliminating the rows with numbers $i$ and~$i+6$ for $i =1, \ldots, 6$. Hence we get the system $S_i = 0$ of polynomial equations in only two variables $\lambda$ and~$\mu$.

\begin{remark}
Due to the form~\eqref{eq:matrixC} of matrix $\mathbf C$, we do not need to compute a $10\times 10$ functional determinant here. Each polynomial~$S_i$ can be found as
\[
\det(\mathbf C_1 + \lambda \mathbf C_2 + \mu \mathbf C_3),
\]
where the $5\times 5$ scalar matrices $\mathbf C_j$ are obtained by a patrial Gauss-Jordan elimination on matrix~$\mathbf C$.
\end{remark}

Let us rewrite the system $S_i = 0$, $i = 1, \ldots, 6$, in form:
\begin{equation}
\label{eq:F0y}
\mathbf F_0\,\mathbf y = \mathbf 0_6,
\end{equation}
where $\mathbf F_0$ is a $6\times 18$ coefficient matrix, and
\begin{multline}
\label{eq:monomials}
\mathbf y =
\left[\lambda^4\mu \quad \lambda^3\mu^2 \quad \lambda^2\mu^3 \quad \lambda\mu^4 \quad \lambda^4 \quad \mu^4 \quad \lambda^3\mu \quad \lambda^2\mu^2 \right.\\ \left. \lambda\mu^3 \quad \lambda^3 \quad \lambda^2\mu \quad \lambda\mu^2 \quad \mu^3 \quad \lambda^2 \quad \lambda\mu \quad \mu^2 \quad \lambda \quad \mu\right]^{\mathrm T}
\end{multline}
is a monomial vector. To solve the system~\eqref{eq:F0y} in a numerically stable way, we perform the following sequence of matrix transformations:
\begin{equation}
\label{eq:seq}
\mathbf F_0 \to \tilde{\mathbf F}_0 \to \mathbf F_1 \to \tilde{\mathbf F}_1 \to \mathbf F_2 \to \tilde{\mathbf F}_2 \to \mathbf F_3 \to \tilde{\mathbf F}_3,
\end{equation}
where each $\tilde{\mathbf F}_i$ is obtained from $\mathbf F_i$ by the Gauss-Jordan elimination with partial pivoting.

The matrix $\mathbf F_1$ of size $8\times 18$ is obtained from $\tilde{\mathbf F}_0$ by adding two new rows: first one corresponds to the last row of $\tilde{\mathbf F}_0$ multiplied by~$\lambda$, second one --- to the next to last row of $\tilde{\mathbf F}_0$ multiplied by~$\mu$.

The matrix $\mathbf F_2$ of size $12\times 18$ is obtained from $\tilde{\mathbf F}_1$ by adding four new rows corresponding to the last two rows of $\tilde{\mathbf F}_1$ multiplied by~$\lambda$ and~$\mu$.

The matrix $\mathbf F_3$ of size $17\times 18$ is obtained from $\tilde{\mathbf F}_2$ by adding five new rows. We multiply the last two rows of $\tilde{\mathbf F}_2$ by $\lambda$ and~$\mu$, and thus get four additional rows. One more row is obtained by multiplying the 10th row of $\tilde{\mathbf F}_2$ by~$\mu$.

Finally we get
\[
\mu = - (\tilde{\mathbf F}_3)_{16, 18}, \quad \lambda = - \mu\,(\tilde{\mathbf F}_3)_{17, 18}.
\]

\begin{remark}
From algebraic point of view, the above sequence~\eqref{eq:seq} \emph{interreduces} the ideal $J = \langle S_i \mid i = 1, \ldots, 6\rangle$. The result is the Gr\"{o}bner basis of~$J$ with respect to the graded lexicographic order. It consists of two polynomials represented by the last two rows of matrix~$\tilde{\mathbf F}_3$.
\end{remark}

Having found $\lambda$ and $\mu$, we compute the entries of $\boldsymbol{\omega}^*$ performing the Gauss-Jordan elimination with partial pivoting on matrix~$\mathbf C$ in~\eqref{eq:matrixC}. Finally, we compute the calibration matrix by the Cholesky decomposition of~$\boldsymbol{\omega}^* = \mathbf K \mathbf K^{\mathrm T}$, and then find (up to scale) the metric camera matrices~$\mathbf P_i^M$ by~\eqref{eq:update}.

\begin{remark}
Note that the matrices $\mathbf R_i$ estimated from~\eqref{eq:metric} are not in general rotations and thus need to be corrected~\cite{Zhang}. We used the singular value decomposition $\mathbf R_i = \mathbf U_i \mathbf D_i \mathbf V_i^{\mathrm T}$ and then replaced~$\mathbf R_i$ by~$\tilde{\mathbf R}_i = \mathbf U_i \mathbf V_i^{\mathrm T}$. It is well-known that the rotation matrix $\tilde{\mathbf R}_i$ is the closest to $\mathbf R_i$ with respect to Frobenius norm.
\end{remark}

\section{Experiments on Synthetic Data}
\label{sec:synth}

\begin{figure}[t]
\centering
\includegraphics[scale=0.3]{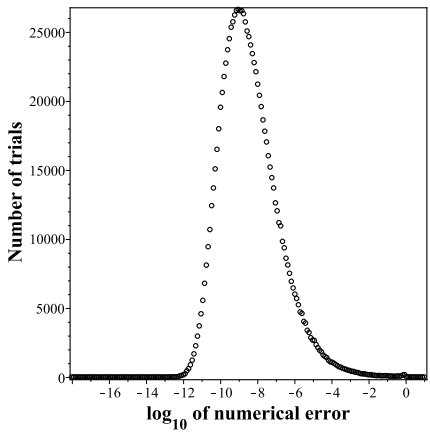}
\caption{Numerical error distribution. Median error is $2.8\times 10^{-9}$.}
\label{fig:numer_err}
\end{figure}

\begin{figure}[t]
\centering
\includegraphics[scale=0.3]{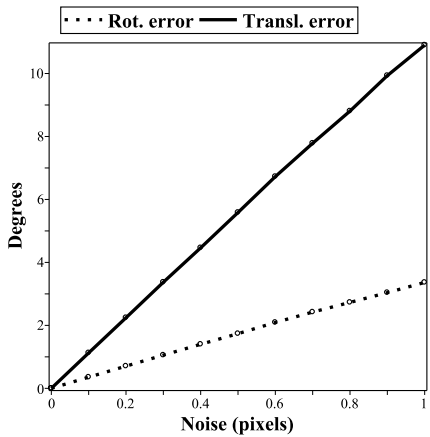}
\caption{Rotational and translational errors relative to noise level.}
\label{fig:transl_err}
\end{figure}

The algorithm has been implemented in C/C++. All computations were performed in double precision. Synthetic data setup is given in Table~\ref{tab:setup}, where the baseline length is the distance between the first and third camera centers. The second camera center varies randomly around the baseline middle point with amplitude~$0.025$.
\begin{table}[ht]
\centering
\begin{tabular}{|c|c|}
\hline
Distance to the scene & 1\\\hline
Scene depth & 0.5\\\hline
Baseline length & 0.1\\\hline
Image dimensions & $352 \times 288$ \\\hline
Calibration matrix & $\begin{bmatrix}425 & 0 & 176 \\ 0 & 425 & 144 \\ 0 & 0 & 1 \end{bmatrix}$ \\\hline
\end{tabular}
\caption{Synthetic data setup.}
\label{tab:setup}
\end{table}

We have measured the numerical error by the value
\[
\frac{\|\mathbf K - \hat{\mathbf K}\|}{\|\hat{\mathbf K}\|},
\]
where $\hat{\mathbf K}$ is the ground truth calibration matrix, $\|\cdot \|$ is the Frobenius norm. The distribution of the numerical error is reported in Figure~\ref{fig:numer_err}, where the total number of trials is $10^6$.

The running time information for our implementation of the algorithm is given in Table~\ref{tab:timing}.
\begin{table}[ht]
\centering
\begin{tabular}{|c|c|c|}
\hline
Step & Projective reconstr. & Metric reconstr.\\\hline
$\mu s$ & 7.9 & 28.4/root\\\hline
\end{tabular}
\caption{Average running times for the algorithm steps on a system with Intel Core i5 2.3 GHz processor.}
\label{tab:timing}
\end{table}

In Figure~\ref{fig:transl_err}, we demonstrate the stability of the algorithm under increasing image noise. We have added a Gaussian noise with a standard deviation varying from 0 to 1 pixel in a $352 \times 288$ image. Each point is a median of $10^6$ trials.

\subsection{Outliers}
\label{ssec:ransac}

To test the algorithm in presence of outliers (incorrect matches), we have modeled a sequence of 70 cameras with centers on a circle, and 400 scene points viewed by all the cameras. For each image, we have added a Gaussian noise with one pixel standard deviation and $20\%$ of outliers (uniformly distributed points in the image plane).

The auto-calibration algorithm was used as a hypothesis generator within a random sample consensus (RANSAC) framework~\cite{FB}. For better computational efficiency we used the \emph{preemptive} RANSAC from~\cite{Nister}. The motion hypotheses were scored by the Sampson approximation to geometric error~\cite{HZ}. The number of hypotheses was set to 400 for each camera position, and the preemption block size was set to~100.

\begin{figure}[t]
\centering
\includegraphics[scale=0.43]{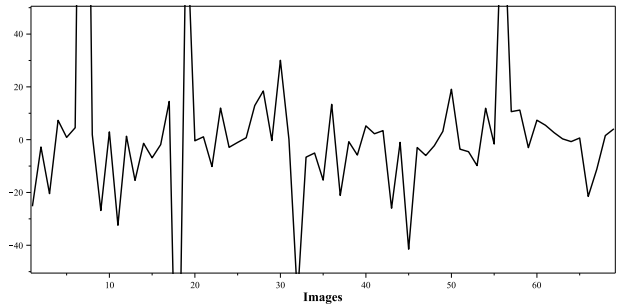}
\caption{Skew parameter $K_{12}$ estimated from the sequence of 70 synthetic images. Average value of $K_{12}$ is $2.16$.}
\label{fig:calib}
\end{figure}

\begin{figure}[t]
\centering
\includegraphics[scale=0.4]{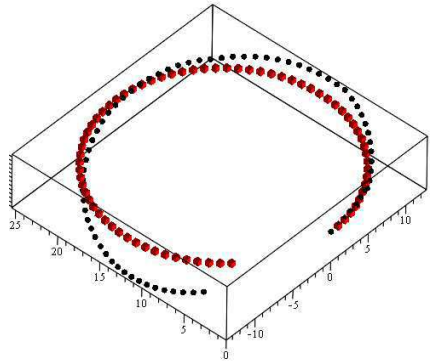}
\caption{The camera track estimated from the sequence of 70 synthetic images. The red solid boxes are the ground truth camera positions.}
\label{fig:track}
\end{figure}

The results are presented in Figure~\ref{fig:calib} and Figure~\ref{fig:track}. No iterative refinements were performed in the estimation. The calibration matrix averaged from the image sequence is as follows:
\[
\mathbf K = \begin{bmatrix}399.52 & 2.16 & 161.54 \\ 0 & 405.37 & 142.14 \\ 0 & 0 & 1 \end{bmatrix}.
\]

\section{Discussion}
\label{sec:concl}

A new non-iterative auto-calibration algorithm is presented. It derives the camera calibration from the smallest possible number of views and scene points. A computation on synthetic data confirms its accuracy and high speed performance. The algorithm is quite flexible. It is reliable, for example, even in case of pure rotations (baseline $= 0$), if the calibration matrix is only needed.

\bibliographystyle{amsplain}

\end{document}